\documentclass[conference, 9pt]{IEEEtran}
\IEEEoverridecommandlockouts

\usepackage{hhline}
\usepackage{colortbl}
\usepackage{centernot}
\usepackage{pbox}
\usepackage{amsmath}
\usepackage{amsfonts}
\usepackage{url}
\usepackage{bm}
\usepackage{comment}
\usepackage{graphicx}
\usepackage[hidelinks]{hyperref}
\usepackage{epstopdf}
\usepackage{multirow}
\usepackage{color}
\usepackage{tabu}
\usepackage{mdframed}
\usepackage{syntax}
\usepackage{fancyvrb}
\usepackage{subfig}
\usepackage{cleveref}
\crefname{figure}{fig.}{figures}
\Crefname{figure}{Fig.}{Figures}
\Crefname{table}{TABLE}{Tables}

\usepackage[ruled,linesnumbered,algo2e]{algorithm2e}
\usepackage[noend]{algpseudocode}
\usepackage{algorithmicx}
\usepackage{algpseudocode}
\usepackage{paralist}
\usepackage{stmaryrd}
\usepackage{tikz}
\usepackage[referable]{threeparttablex}
\usepackage{tabularx}
\usepackage{booktabs}
\usepackage{array}
\usepackage{fancyhdr}
\usepackage{float}
\usepackage{xspace}
\usepackage{pifont}
\usepackage{balance}
\usepackage{xcolor}

\usepackage{tablefootnote}
\usepackage[normalem]{ulem}

\usepackage{tikz}
\usepackage{xcolor}
\newcommand*\circled[1]{\tikz[baseline=(char.base)]{
            \node[shape=circle,fill,inner sep=1pt,scale=0.8] (char) {\textcolor{white}{#1}};}}

\usepackage{graphicx}
\usepackage{adjustbox}

\usepackage{float}

\newfloat{figtab}{htb}{fgtb}
\makeatletter
  \newcommand\figcaption{\def\@captype{figure}\caption}
  \newcommand\tabcaption{\def\@captype{table}\caption}
\makeatother

\definecolor{princetonorange}{RGB}{255,143,0}

\definecolor{lightgreen}{RGB}{198, 224, 183}

\definecolor{lightred}{RGB}{240, 205, 176}

\begin{document}

\title{NetEncoder: General Netlist Representation Learning via Self-Supervised Text-Attributed Graph}
\title{Towards Netlist Foundation Model: A Self-Supervised and Cross-Stage-Aware Netlist Encoder via Text-Attributed Graph}
\title{Towards Netlist Foundation Model: A Multimodal Cross-Stage-Aligned Netlist Encoder via Text-Attributed Graph}
\title{Towards Netlist Foundation Model: \\ A Multimodal and Cross-Stage-Aligned Netlist Encoder \\ via Text-Attributed Graph}
\title{NetTAG: A Multimodal and Cross-Stage-Aligned \\ Netlist Foundation Model via Text-Attributed Graph}

\title{NetEncoder: A Multimodal RTL-and-Layout-Aligned \\ Netlist Foundation Model via Text-Attributed Graph}

\title{NetTAG: A Multimodal RTL-and-Layout-Aligned \\ \underline{Net}list Foundation Model via \underline{T}ext-\underline{A}ttributed \underline{G}raph}

\title{NetTAG-Gen: Unleashing Generative Capability of Netlist Foundation Model via Encoder-Decoder Alignment}

\title{NetTAG-Gen: A Generative Netlist Foundation Model via Multimodal Encoder-Decoder Alignment}


\title{GenEDA: Unleashing \underline{Gen}erative Reasoning on Netlist via Cross-Modal \underline{E}ncoder-\underline{D}ecoder-\underline{A}ligned Foundation Model\looseness=-1}

\title{GenEDA: Towards \underline{Gen}erative Netlist Functional Reasoning \\ via Cross-Modal Circuit \underline{E}ncoder-\underline{D}ecoder \underline{A}lignment\looseness=-1}






\author{
    \IEEEauthorblockN{
        Wenji Fang,
        Wang Jing,
        Yao Lu,
        Shang Liu,
        Zhiyao Xie$\textsuperscript{*}$
    }
    \IEEEauthorblockA{Hong Kong University of Science and Technology}
    {$\textsuperscript{*}$Corresponding Author\vspace{-.1in}}
}

\maketitle


\begin{abstract}

The success of foundation AI has motivated the research of circuit foundation models, which are customized to assist the integrated circuit (IC) design process. However, existing pre-trained circuit foundation models are typically limited to standalone encoders for predictive tasks or decoders for generative tasks. These two model types are developed independently, operate on different circuit modalities, and reside in separate latent spaces. This restricts their ability to complement each other for more advanced capabilities. In this work, we present GenEDA, the first framework that cross-modally aligns circuit encoders with decoders within a shared latent space. GenEDA bridges the gap between graph-based circuit representation learning and text-based large language models (LLMs), enabling communication between their respective latent spaces. To achieve the alignment, we propose two paradigms to support both open-source trainable LLMs and commercial frozen LLMs. We leverage this aligned architecture to develop the first generative foundation model for netlists, unleashing LLMs’ generative reasoning capability on the low-level and bit-blasted netlists. GenEDA enables three unprecedented generative netlist functional reasoning tasks, where it reversely generates high-level functionalities such as specifications and RTL code from low-level netlists. These tasks move beyond traditional gate function classification to direct generation of full-circuit functionality. Experiments demonstrate that GenEDA significantly boosts advanced LLMs' (e.g., GPT and DeepSeek series) performance in all tasks.\looseness=-1 


\end{abstract}


\maketitle

\section{Introduction}
\label{sec:intro}


The ever-increasing IC complexity and skyrocketing IC design costs are challenging traditional electronic design automation (EDA) techniques. This trend has motivated the community's active exploration of new IC design methods, such as AI-assisted EDA techniques. Most recently, emerging \emph{foundation AI models} have been customized and applied to IC design, named circuit foundation models~\cite{chen2024large, fang2025cfm}. These pre-trained circuit foundation models target more generalized AI solutions for IC design.







\begin{figure}[!t]
  \centering
  \includegraphics[width=1\linewidth]{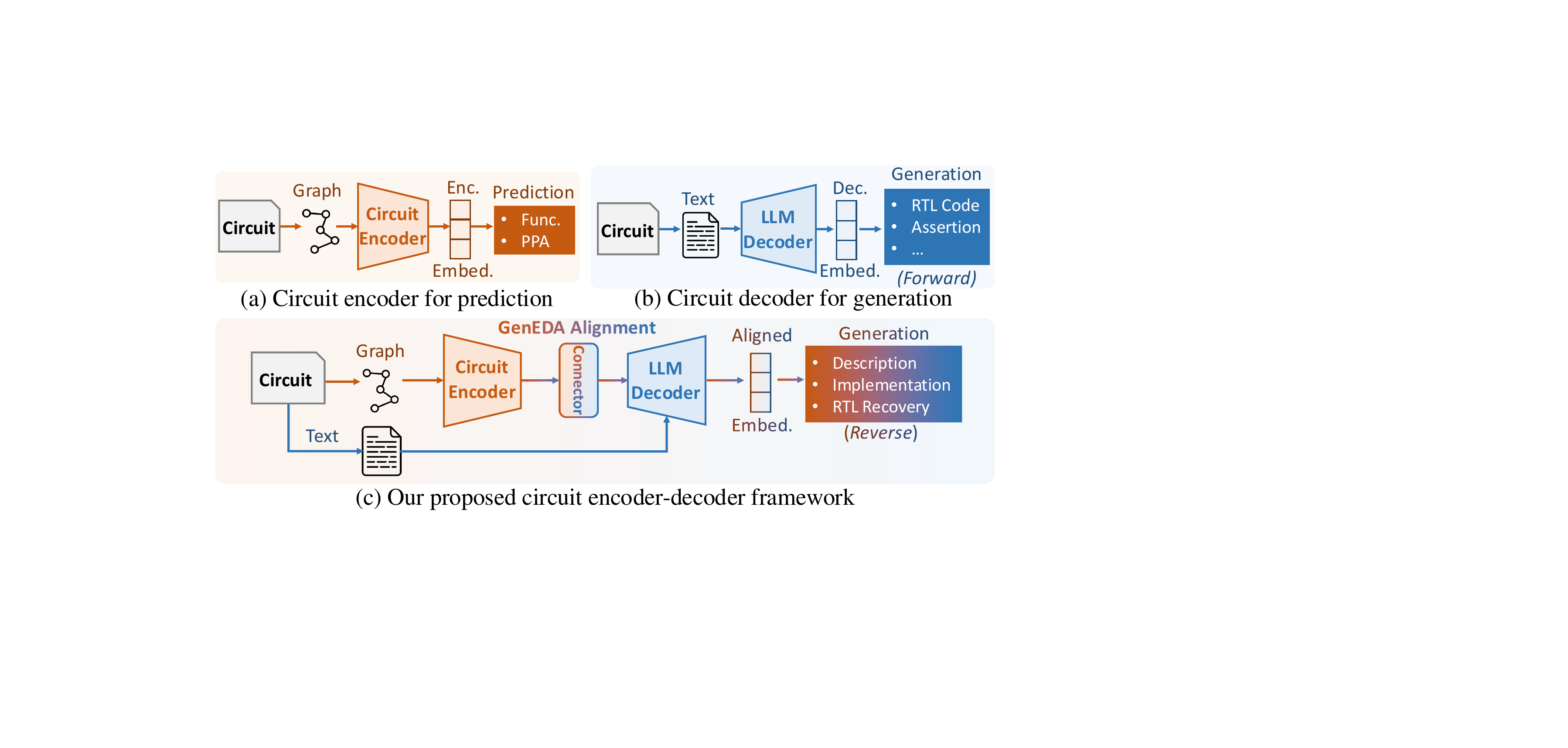}
\vspace{-.2in}
  \caption{(a) \& (b) Two main types of existing circuit foundation models: encoders for prediction and decoders for generation. (c) \emph{GenEDA} proposes a general framework that aligns circuit \emph{encoders} with \emph{decoders}, leveraging encoder-captured information to enhance decoder generation.} 
  \label{fig:paradigm}
    \vspace{-.2in}
\end{figure}

\textbf{Circuit foundation model: two main types.}  
As \Cref{fig:paradigm} shows, existing circuit foundation models can be categorized into two main types: (a) circuit encoder for predictive tasks, and (b) circuit decoder for generative tasks. 
\Cref{tbl:works} provides a detailed comparison of these two types of works. 
\textbf{Circuit encoder} refers to pre-trained models that encode circuits into general embeddings (i.e., circuit representation learning).  
Taking these embeddings as input, lightweight downstream models are then fine-tuned for various \textit{predictive} EDA tasks, such as circuit functional reasoning~\cite{fang2025circuitencoder, shi2024deepgate3, shi2023deepgate2, li2022deepgate, wang2022functionality} and circuit quality prediction~\cite{fang2025circuitfusion, wu2025circuit, deng2024less, xu2023fast}. \textbf{Circuit decoder} refers to pre-trained models with circuit-related \textit{generative} capability. Built mostly on large language models (LLMs), these decoders generate text outputs, such as circuits in RTL code~\cite{zhang2024mg, pei2024betterv, liu2023rtlcoder}, verification assertions~\cite{fang2024assertllm, kande2024security}, EDA tool scripts~\cite{he2023chateda}, etc.  



\textbf{Limitation: lack of alignment.} 
Developed independently, circuit encoders and decoders operate on different modalities and handle circuit representations in clearly \textit{distinct latent spaces}. 
Specifically:
(a) Circuit encoders typically work in the circuit graph latent space. They excel at capturing circuit structural and functional properties into embeddings for predictive tasks, but lack generative capabilities. (b) Circuit decoders, typically LLMs, operate in the text latent space. They are effective at generating circuit-related text (e.g., RTL code, assertions), but they rely solely on textual input and cannot fully utilize the underlying structural information of circuits. 
As a result, these two important types are not aligned due to an inherent gap in their latent space, preventing more advanced capabilities. 
\looseness=-1

\textbf{GenEDA: an encoder-decoder alignment framework.} 
To address this, we present \emph{GenEDA}\footnote{The code of GenEDA and benchmarks of proposed tasks are available at: https://github.com/hkust-zhiyao/GenEDA}, the first framework that aligns pre-trained circuit encoders with LLM decoders. 
GenEDA bridges these two major types by communicating \emph{circuit graph latent space} and \emph{text latent space}, enabling effective information exchange. This alignment allows structural and functional insights captured by circuit encoders to directly enhance the generative capabilities of LLM decoders.\looseness=-1 


GenEDA achieves the alignment by proposing two paradigms for both open-source trainable LLMs and commercial frozen LLMs: (1) Embedding-based alignment, which fine-tunes trainable LLMs using graph-based circuit embeddings by introducing a modality \emph{connector}, and (2) Prediction-based alignment, which augments commercial frozen LLMs by feeding them textual predictions from the encoder.\looseness=-1

\begin{table*}[!t]
\center
\vspace{-.2in}
\resizebox{0.8\textwidth}{!}{
\begin{tabular}{c|c|cc|cccc} \toprule
                                        &                                                                                                        & \multicolumn{2}{c|}{\textbf{Input}}                                                                                         & \multicolumn{4}{c}{\textbf{Task}}                                                                                                                    \\ \cline{3-4} \cline{5-8}
\multirow{-2}{*}{\textbf{Type}} & \multirow{-2}{*}{\textbf{Method$^\dagger$}}                                                                      & \footnotesize Format                          & \footnotesize Modality                                                               & \footnotesize Pred.                     & \footnotesize Gen.                      & \footnotesize Description                           & \footnotesize Direction$^\ddagger$        \\ \midrule \midrule
                                        & \footnotesize DeepGate~\cite{zheng2025deepgate4}, etc. & AIG                                      & Graph                                                                           & \checkmark                                  &                                    & function pred.                                 & Reverse                   \\ \cline{3-4}
                                        & \footnotesize CircuitFusion~\cite{fang2025circuitfusion}                                                                                          & RTL                                      &                                                                                 & \checkmark                                   &                                    & quality pred.                                  & Forward                   \\
                                        & \footnotesize MGVGA~\cite{wu2025circuit}                                                                                                  & AIG                                 &                                                                                 & \checkmark                                   &                                    & func./quality pred.                            & Reverse/Forward                       \\
\multirow{-4}{*}{Encoder}      & \footnotesize NetTAG~\cite{fang2025nettag}                                                                                                 & Netlist                                  & \multirow{-3}{*}{\begin{tabular}[c]{@{}c@{}}Graph \\      \& Text\end{tabular}} &  \checkmark                                  &                                    & func./quality pred.                            & Reverse/Forward                       \\ \midrule
                                        & \footnotesize RTLCoder~\cite{liu2024rtlcoder}, etc.                         & Spec                                     &                                                                                 &                                    &  \checkmark                                  & RTL gen.                                       &    Forward                        \\
                                        & \footnotesize HDLDebugger~\cite{yao2024hdldebugger}, etc.                 & RTL                                      &                                                                                 &                                    & \checkmark                                    &  RTL debug                                      &    Forward                       \\
                                        & \footnotesize AssertLLM~\cite{fang2024assertllm}, etc                    & Spec                                     & \multirow{-3}{*}{\begin{tabular}[c]{@{}c@{}}Text \\      (Image)\end{tabular}}  &                                    &  \checkmark                                  & assertion gen.                                 & Forward \\ \cline{3-4}
\multirow{-4}{*}{Decoder}      & \footnotesize DeepRTL~\cite{liu2025deeprtl}$^\star$                                                                                                & RTL/Spec                                 & Text                                                                            &                                    & \checkmark                & RTL understand./gen.                         & Reverse/Forward                       \\ \midrule
\cellcolor[HTML]{DAE9F8}\textbf{Enc-Dec}                        & \cellcolor[HTML]{DAE9F8}\textbf{GenEDA (ours)}                                                                & \cellcolor[HTML]{DAE9F8}\textbf{Netlist} & \cellcolor[HTML]{DAE9F8}\textbf{Graph\&Text}                                  & \cellcolor[HTML]{DAE9F8}\textbf{\checkmark} & \cellcolor[HTML]{DAE9F8}\textbf{\checkmark} & \cellcolor[HTML]{DAE9F8}\textbf{Function gen.} & \cellcolor[HTML]{DAE9F8}\textbf{Reverse}          \\ \bottomrule         
\end{tabular}

}
\begin{tablenotes}\footnotesize
\item $^\dagger$ We list one of the representative works for each method in the table. For a more comprehensive comparison, please refer to~\Cref{sec:work-method}.
\item $^\ddagger$ Task direction is \textit{forward} if they follow the VLSI design flow (e.g., predicting quality at the early stage, generating RTL from spec), and \textit{reverse} if they go against it (e.g., predicting or generating function from netlist). Reverse tasks are challenging due to the design flow’s irreversible nature.
\item $^\star$ This work~\cite{liu2025deeprtl} leverages T5, an encoder-decoder LLM. However, it targets only generative tasks, so we categorize it as a circuit decoder.
\end{tablenotes} 
\vspace{-.05in}
\caption{Comparison of GenEDA with representative categories of circuit foundation models. Existing circuit encoders mainly leverage graph structure for prediction tasks, while circuit decoders focus on semantic text for generation tasks. GenEDA bridges the gap between these two widely explored branches by aligning encoders and decoders within a shared latent space. After alignment, GenEDA supports more challenging generative netlist functional reasoning tasks.}
\vspace{-.2in}
\label{tbl:works}
\end{table*}

\textbf{Challenging application: generative functional reasoning of netlist.}
GenEDA's alignment framework can support unprecedentedly challenging generative applications, such as \textit{generative} reasoning of the \textit{netlist} functionality\footnote{In the reverse netlist functional reasoning scenario, the high-level specification and RTL code as ground-truth are unknown to the model. Models are only provided with the low-level netlist as inputs.}. 
Unlike RTL code, netlists are composed of a huge number of low-level, bit-blasted gates and their complex connections, offering little human-readable semantics for LLMs to understand.
While prior netlist encoders~\cite{wang2022functionality, wu2023gamora, wang2024fgnn2, deng2024less, fang2025nettag} can extract structural and functional features into embeddings, they are limited to predictive reasoning, only classifying the functionality of individual gates. GenEDA bridges this gap by aligning the encoder’s structural and functional understanding of netlists with the decoder’s generative strengths. This encoder-decoder alignment enables generating high-level functionality directly from low-level netlist inputs, which is an unprecedentedly challenging task due to the irreversible nature of logic synthesis.

Specifically, GenEDA reasons functionalities of given netlists in a wide spectrum of granularities, with outputs including: (1) general function description, (2) circuit implementation details, and (3) fine-grained exact RTL code. 
These GenEDA-supported new generative netlist functional reasoning tasks are highly valuable in multiple aspects: \textbf{(1) Practical applications:} Reasoning high-level functionality from bit-blasted netlists can support critical applications~\cite{kuehlmann2002robust, subramanyan2013reverse,wu2023gamora}, such as functional verification, datapath optimization, and malicious logic detection.
\textbf{(2) Unprecedented reasoning quality:} These tasks move beyond traditional gate function classification by directly generating the overall functionality of entire circuits, including specifications and RTL code, offering a significant leap in reasoning quality.
\textbf{(3) Benchmarking model capability:} Our proposed tasks introduce new benchmarks for evaluating the generative circuit reasoning capabilities and netlist understanding of foundation models. Since these tasks generate human-readable circuit information, they help enhance the interpretability of circuit models.\looseness=-1

The contributions of this paper are summarized as follows:
\begin{itemize}
    \item \textbf{Circuit encoder-decoder alignment framework.} We propose the first framework that cross-modally aligns pre-trained circuit encoders with LLM decoders. It supports both trainable and frozen LLMs for advanced generative tasks through two alignment paradigms.
    \item \textbf{Generative netlist foundation model.} Leveraging this framework, we develop the first generative foundation model for netlists. By integrating structural and functional insights captured by netlist encoders, the model unleashes LLMs' generative reasoning capability on the low-level and bit-blasted netlists.
    \item \textbf{New generative netlist reasoning tasks and benchmarks.} We propose three novel generative netlist functional reasoning tasks with corresponding benchmarks, advancing beyond prior gate function classification. We also release these benchmarks to encourage follow-up research on these tasks.
    \item \textbf{Boosting SOTA LLMs' performance.} Experimental results validate that GenEDA significantly boosts the performance of cutting-edge LLMs on all three new functional reasoning tasks after alignment with the pre-trained netlist encoder.
\end{itemize}

\section{Related Work}

\begin{figure}[!b]
  \centering
  \includegraphics[width=1\linewidth]{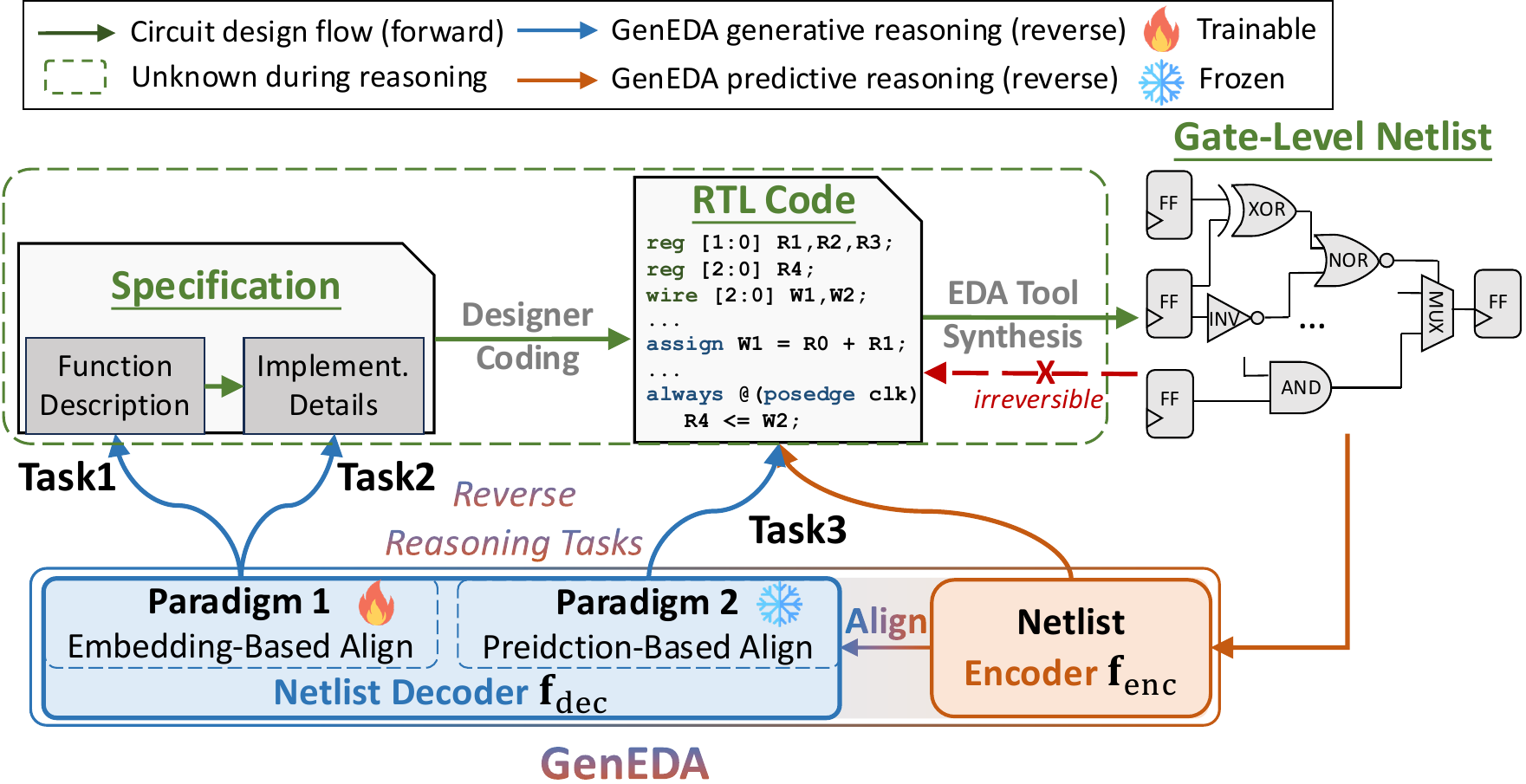}
  \vspace{-.2in}
  \caption{Overview of GenEDA framework integrated into the standard IC design flow. GenEDA aligns the pre-trained netlist encoder with LLM decoders through two alignment paradigms, enabling challenging generative netlist functional reasoning tasks.} 
  \label{fig:overview}
  \vspace{-.2in}
\end{figure}

\begin{figure*}[!t]
  \centering
  \vspace{-.3in}
  \includegraphics[width=1\linewidth]{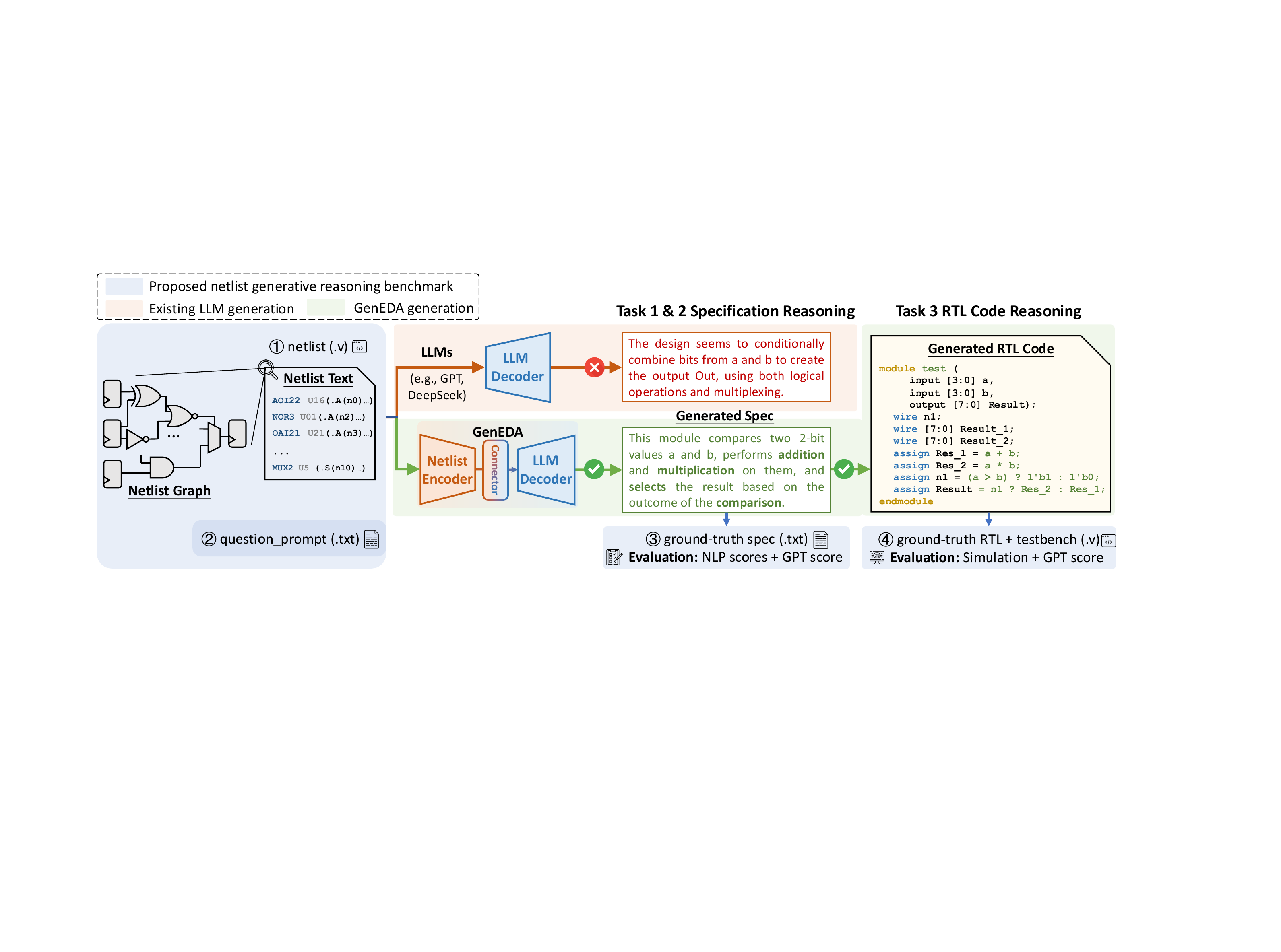}
  \vspace{-.22in}
  \caption{Proposed generative netlist functional reasoning benchmarks. For Tasks 1 and 2, the netlist and question prompt are processed by models for specification reasoning, and evaluated using NLP scores and GPT scores. For Task 3, the models reconstruct RTL code from the netlist, which is evaluated via simulation and GPT scores.} 
  \label{fig:task}
  \vspace{-.2in}
\end{figure*}

\subsection{Method Related: Circuit Foundation Model}
\label{sec:work-method}

Recent advances in foundation AI for EDA have enabled strong generalization and generation capabilities through the \emph{pretrain-finetune} process. As summarized in~\Cref{tbl:works}, these circuit foundation models can be categorized into encoder-based and decoder-based architectures, each supporting different inputs and tasks.

\textbf{Circuit encoders for prediction.}
Encoder-based models typically learn structured circuit representations to support predictive tasks such as reverse functional reasoning and early-stage design quality estimation. Most methods~\cite{li2022deepgate, shi2023deepgate2, shi2024deepgate3, khan2025deepseq2, zheng2025deepgate4, wang2022functionality, wang2024fgnn2, deng2024less, PolarGate} focus on AIG netlists and use graph learning to capture the circuit structure. Recent work~\cite{fang2025circuitfusion, wu2025circuit, fang2025nettag} fuses multiple modalities (graph, text, image), but their multimodal fusion are limited to the encoder side, lacking alignment with LLMs for enabling generation capabilities.
\textbf{Circuit decoders for generation.}
LLM-based decoders support forward-generation tasks like RTL or assertion generation~\cite{liu2023chipnemo, liu2024rtlcoder, chang2023chipgpt, fang2024assertllm, sun2023towards}, debugging~\cite{yao2024hdldebugger}, optimization~\cite{pei2024betterv, yao2024rtlrewriter}, and knowledge querying~\cite{jiang2024fabgpt, wu2024chateda}. DeepRTL~\cite{liu2025deeprtl} further explores bidirectional generation between RTL and specification. However, since RTL is easier for LLMs to understand due to its rich semantics, the text-only T5 LLM is sufficient for these tasks. Additionally, existing multimodal approaches focus on circuit images~\cite{yao2024rtlrewriter, fang2024assertllm}, a capabiltity already present in current multimodal LLMs. They overlook the crucial circuit graph structure, which remains largely unexplored in combination with LLMs.

\textbf{Circuit encoder-decoder alignment by GenEDA.}
GenEDA bridges this gap by proposing the first cross-modal encoder-decoder alignment framework, which aligns circuit structural representations with LLM textual representations in a shared latent space, enabling support for advanced generative tasks.\looseness=-1

\subsection{Application Related: Functional Reasoning on Netlists}
Functional reasoning on gate-level netlists aims to reconstruct the high-level functionality originally described in specifications or RTL.
It plays a critical role in functional verification, logic optimization, datapath synthesis, and hardware security. Existing approaches primarily fall into two categories: formal methods for analyzing functionality~\cite{kuehlmann2002robust,li2012reverse, subramanyan2013reverse, gascon2014template}, and machine learning methods for gate-level function classification~\cite{alrahis2021gnn, chowdhury2021reignn, wang2022functionality, wu2023gamora, deng2024less}. We detail these two categories below.


\textbf{Formal analysis.}
Traditionally, netlist functional reasoning relies on structural and functional analysis using formal techniques~\cite{kuehlmann2002robust, li2012reverse, subramanyan2013reverse, gascon2014template}. These approaches typically extract subcircuits from the netlist and match them against components in a golden library via exhaustive formal verification. However, they are time-consuming, dependent on library completeness, and incapable of recognizing functional variants.
\textbf{GNN-based individual gate function \textit{prediction}.}
Recently, GNN-based methods~\cite{alrahis2021gnn, chowdhury2021reignn, wang2022functionality, wu2023gamora, deng2024less} have been applied to predictive netlist functional reasoning. These methods focus on learning the netlist structure for gate-level function classification.
While they effectively identify the roles of known components, they are limited in generalizing to unseen functionality and cannot reason about the full behavior of the entire circuit.
\looseness=-1

\textbf{GenEDA-based entire circuit function \textit{generation}.}
GenEDA advances from gate-level prediction to full-circuit generative reasoning. By aligning netlist encoders with LLM decoders, it enables the direct generation of high-level specifications and RTL code from low-level netlists.\looseness=-1 

\section{Overview}




\Cref{fig:overview} presents the overview of our GenEDA framework, integrated into the standard digital IC design flow. GenEDA aligns the state-of-the-art post-synthesis netlist encoder NetTAG~\cite{fang2025nettag} $\mathbf{f}_\text{enc}$ with cutting-edge LLM-based decoders $\mathbf{f}_\text{dec}$.
It first converts the circuit netlist $\mathcal{N}$ into embeddings via the encoder $\mathbf{f}_\text{enc}$, capturing both netlist structural and functional information. The encoder output is then provided to aligned decoders $\mathbf{f}_\text{dec}$ to support advanced generative reasoning tasks on netlists.\looseness=-1

We propose two novel encoder-decoder alignment paradigms for both open-source trainable LLMs and commercial frozen LLMs: (1) Embedding-based alignment, where a trainable LLM decoder is instruction-tuned with encoder embeddings with a circuit modality connector. (2) Prediction-based alignment, where the encoder annotates textual gate function predictions on netlists to augment the inputs to the frozen LLM.
In our experiments, we apply paradigm 1 for specification reasoning tasks (Task 1 and Task 2) and paradigm 2 for exact RTL code reasoning (Task 3).
\looseness=-1




The rest of the paper is organized as follows:
In~\Cref{sec:task}, we first provide a detailed explanation of our three novel generative netlist functional reasoning tasks and our contributed benchmarks. 
In~\Cref{sec:align}, we illustrate the two proposed encoder-decoder alignment paradigms, as well as how to tackle the three tasks.
In~\Cref{sec:expr}, we demonstrate the effectiveness of GenEDA in all three generative functional reasoning tasks in experiments. 
Finally, in~\Cref{sec:discussion}, we discuss the potential of extending GenEDA to other circuit design stages and the broader applications of generative netlist functional reasoning.\looseness=-1

\section{Benchmarking Generative Functional Reasoning}
\label{sec:task}
\Cref{fig:task} illustrates our three novel generative functional reasoning tasks on netlists, along with the benchmarks developed to support them. 
These tasks aim to reversely generate high-level circuit functionality, including natural language specifications and exact RTL code, from low-level bit-blasted netlists. Please note that models are only provided with netlists, their corresponding specifications and RTL code serve as ground-truths and and unknown to the model.
Our proposed benchmarks evaluate the generative model’s ability to truly understand the functionality of netlists, setting a new direction for generative EDA tasks. We detail the three tasks below.\looseness=-1

\begin{figure*}[!t]
  \centering
  \vspace{-.3in}
  \includegraphics[width=1\linewidth]{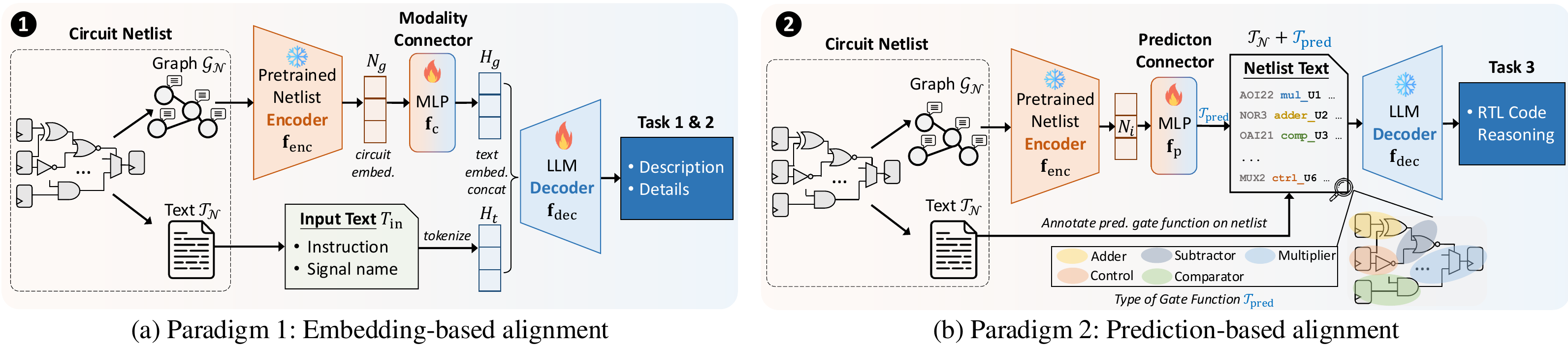}
  \vspace{-.2in}
  \caption{Two encoder-decoder alignment paradigms in GenEDA: 1. Embedding-based alignment integrating embeddings from the netlist encoder with trainable LLM decoders, and 2. Prediction-based alignment using functional predictions from the netlist encoder as textual input for frozen LLM decoders.} 
  \label{fig:align}
  \vspace{-.15in}
\end{figure*}

\subsection{Task 1 \& 2: specification reasoning from netlist.}

\textbf{Task and benchmark description.}
Tasks 1 and 2 aim to reversely generate high-level natural language specifications from gate-level netlists, as shown in~\Cref{fig:task}.
\textbf{Task 1 Function description generation}  focuses on generating circuit functional descriptions from low-level netlists, emphasizing the overall behavior of the circuit.
\textbf{Task 2 Implementation detail generation} targets the reconstruction of step-by-step signal propagation and logic behavior from the netlist, reflecting the underlying design implementation.
As these are novel tasks with no prior benchmarks, we construct new datasets and benchmarks to support model training and evaluation. Specifically, we collect 400 circuit netlists with various design
scales and complexities, annotated with natural language specifications as ground-truth. For each design, our proposed benchmark provides the following information in three separate files:
\begin{itemize}
    \item \textbf{Netlist text.} Gate-level netlist in Verilog text format synthesized from RTL code. Please note that in these reverse tasks, the RTL code is unknown to the model. 
    \item \textbf{Question prompt.} For Task 1, the prompt inquires models to describe the interface, purpose, functionality, and constraints of the netlist. For Task 2, the prompt asks the model to explain the combinational logic, sequential behavior, and control flow.
    \item \textbf{Ground-truth specification text.} Since real-world RTL specifications are rarely available, following~\cite{liu2025deeprtl, fang2025circuitfusion}, we generate ground-truth reference answers for these two tasks using GPT-4o prompted with RTL code and manually verify their quality through expert review.\looseness=-1
\end{itemize}

\textbf{Evaluation metrics.}
As there are no standard metrics for evaluating the functional similarity of specification texts, we follow prior work DeepRTL~\cite{liu2025deeprtl} (i.e., generating specification from RTL code) to adopt a combination of natural language metrics. Specifically, we use BLEU, ROUGE-1/2/L, and cosine similarity computed from text embeddings. Additionally, GPT-4o is employed as an automated evaluator to assess semantic similarity between generated outputs and reference specifications. The prompt templates used are provided in~\Cref{sec:expr}.

\subsection{Task 3: arithmetic RTL code reasoning from netlist.}
\textbf{Task and benchmark description.}
\textbf{Task 3 arithmetic RTL code reasoning} targets the reverse generation of RTL code from gate-level netlists, specifically for arithmetic circuits, as shown in~\Cref{fig:task}. Unlike existing \textit{forward} RTL code generation benchmarks~\cite{lu2024rtllm, liu2023verilogeval}, which rely on human-readable specifications as input, our \textit{reverse} task begins with post-synthesis netlists. This makes the task significantly more challenging due to the irreversible nature of logic synthesis and the lack of high-level functional information.

We propose the first benchmark for reverse RTL code reasoning. Given the task’s complexity, we begin by focusing on arithmetic modules. Specifically, we extend the GNN-RE gate-level arithmetic function prediction dataset~\cite{alrahis2021gnn} into a generative benchmark, incorporating golden RTL code and testbenches. For each of the 9 arithmetic designs in our benchmark, we provide:
\begin{itemize}
    \item \textbf{Netlist text.} Gate-level netlist from GNN-RE dataset~\cite{alrahis2021gnn}, originally used for gate function prediction task. 
    \item \textbf{Question prompt.} Instructions to first infer the word-level arithmetic function, and then implement it using RTL code.
    \item \textbf{Ground-truth RTL code.} The original RTL design used for synthesis, which serves as the reconstruction target.
    \item \textbf{Testbench.} A verified testbench with pre-defined module name and IO ports, containing multiple input-output cases for functional validation.
\end{itemize}

\textbf{Evaluation metrics.}
Similar to the existing forward RTL generation evaluation metrics~\cite{lu2024rtllm, liu2023verilogeval}, we validate both syntax correctness and function correctness of the reversely generated arithmetic RTL code using our provided golden testbenches. Additionally, GPT-4o is employed to assign a function similarity score, measuring how closely the generated RTL matches the ground truth. We provide detailed evaluation method implementation and prompts for obtaining GPT-score in~\Cref{sec:expr}.

\section{GenEDA Encoder-Decoder Framework}\label{sec:align}

\Cref{fig:align} presents our proposed two paradigms for the encoder-decoder alignment, accommodating both trainable open-source LLMs and frozen commercial LLMs. 
Through alignment, GenEDA integrates rich netlist information captured by the encoder to enhance the generative reasoning capabilities of decoder LLMs. These two paradigms with their supported tasks are introduced in Section \ref{subec:paradigm1} and \ref{subec:paradigm2}, respectively. In our experiments, Paradigm 1 is applied to Tasks 1 and 2 for specification reasoning, while Paradigm 2 is used for Task 3 to generate exact RTL code.
\looseness=-1

\textbf{Netlist encoder and LLM decoder in GenEDA.} 
Before discussing the alignment mechanisms, we briefly introduce the models used in GenEDA. For the encoder, we employ NetTAG~\cite{fang2025nettag}, the state-of-the-art encoder capable of handling post-synthesis netlists, whereas most prior netlist encoders are limited to the AIG format. NetTAG introduces a text-attributed graph format for netlists and employs a multimodal architecture: it encodes the gate function via symbolic Boolean expressions using an LLM encoder and captures global circuit structure through a graph transformer. This results in rich embeddings that encode both functional and structural information.
On the decoder side, we consider both general-purpose LLMs, such as OpenAI GPT and DeepSeek, and circuit-specific LLMs like RTLCoder~\cite{liu2024rtlcoder}, which is fine-tuned for forward RTL code generation.\looseness=-1

\subsection{Paradigm 1: Embedding-Based Alignment}\label{subec:paradigm1}

\textbf{Paradigm \circled{1} overview.} \Cref{fig:align} (a) shows our paradigm 1, which aligns trainable open-source LLM decoders $\mathbf{f}_\text{dec}$ with the embeddings generated by the netlist encoder $\mathbf{f}_\text{enc}$ from the netlist graph (i.e., text-attributed graph) modality. 
The main challenge is the modality gap between the encoder-generated netlist embeddings and the text embeddings expected by the decoder, as they lie in fundamentally different latent spaces.
To address this, our paradigm 1 introduces a modality connector $\mathbf{f}_\text{c}$, which acts as a ``circuit tokenizer'', transforming the circuit embeddings from our encoder into text-alike embeddings compatible with the LLM decoder. The connector $\mathbf{f}_\text{c}$ is cross-modally instruction-tuned with the LLM $\mathbf{f}_\text{dec}$, enabling deep embedding-level alignment in the shared latent space.
We describe our aligned model architecture and training method below.\looseness=-1

\begin{figure}[!t]
  \centering
  \includegraphics[width=1\linewidth]{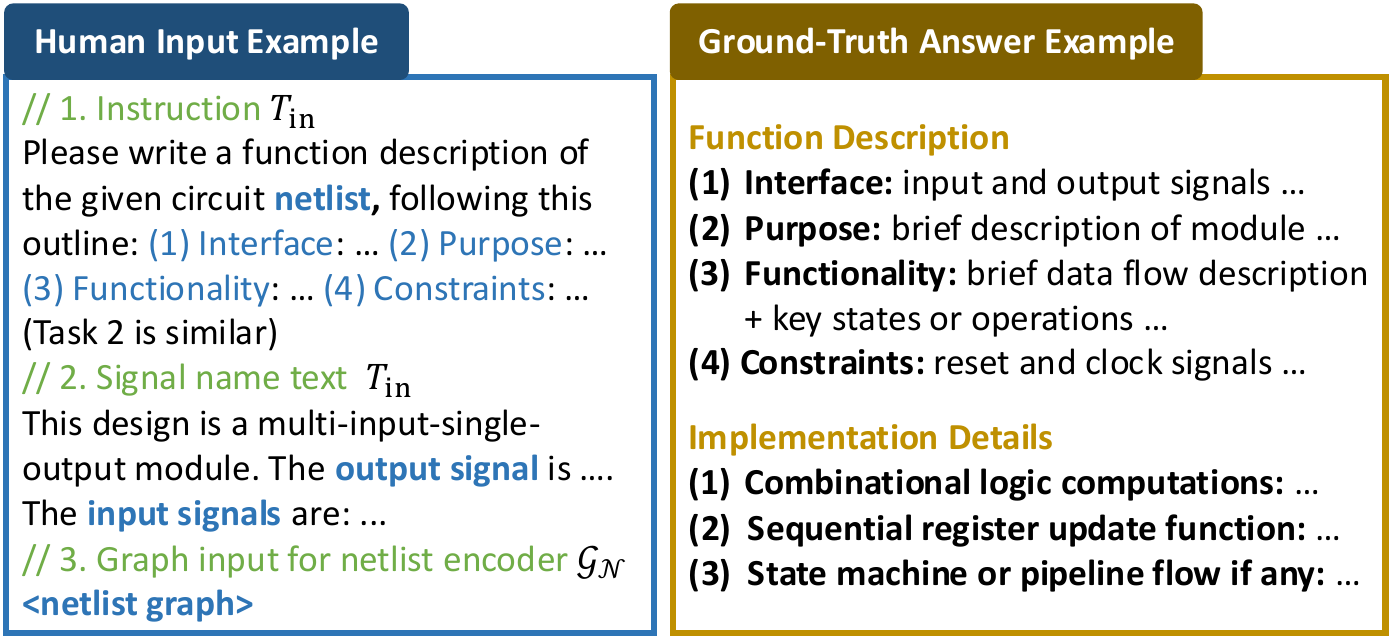}
  \vspace{-.2in}
  \caption{Instruction tuning data pair of alignment paradigm 1.} 
  \label{fig:gpt}
  \vspace{-.15in}
\end{figure}

\textbf{Embedding alignment architecture.}
As shown in~\Cref{fig:align} (a), given an input netlist $\mathcal{N}$, we represent it in two modalities: text-attributed graph $\mathcal{G}_\mathcal{N}$ and netlist Verilog code $\mathcal{T}_\mathcal{N}$ in text format.
For $\mathcal{G}_\mathcal{N}$, we employ our pre-trained netlist encoder $\mathbf{f}_\text{enc}$ to process the netlist and generate the netlist graph-level embedding $N_{\text{g}}$, which contains the structural and functional netlist information. For the netlist Verilog text $\mathcal{T}_\mathcal{N}$, we keep only the signal names and leave the structural details to be captured by the encoder model. 
As shown in~\Cref{fig:gpt}, the signal name text is combined with the question prompt instruction to form the textual input $\mathcal{T}_\text{in}$. This input $\mathcal{T}_\text{in}$ is then tokenized into language embeddings $H_{\text{t}}$ and fed into the LLM decoder $\mathbf{f}_\text{dec}$.

To align the encoded embeddings $N_{\text{g}}$
with the decoder $\mathbf{f}_\text{dec}$ embedding $H_{\text{t}}$, we introduce a trainable connector MLP $\mathbf{f}_c$ to transform $N_{\text{g}}$ into language-modal embedding tokens $H_g$.  These converted embeddings have the same dimension as the word embedding (i.e., $H_{\text{t}}$) space in the trainable LLM decoder $\mathbf{f}_\text{dec}$:
\begin{equation}
\begin{split}
\mathbf{f}_\text{dec}(H_g, H_t) &\Rightarrow \text{Generative Reasoning}, \\ \text{ with }
H_g = \mathbf{f}_\text{c}(N_\text{g}) &\text{ and }  N_\text{g} = \mathbf{f}_\text{enc}(\mathcal{G}_\mathcal{N}).
\end{split}
\end{equation}

\textbf{Training for encoder-decoder alignment.}
To enhance embedding alignment, we propose cross-modal instruction tuning using task-specific datasets (i.e., from Task 1 \& 2) to achieve encoder-decoder alignment. 
As shown in~\Cref{fig:gpt}, we generate a multimodal instruction-response pair for each netlist, where the input includes: (1) A task-specific instruction (e.g., request for function description or implementation details). (2) Input and output signals extracted from the netlist code. (3) The netlist graph format, with the encoder capturing the structural and functional information. The ground truth is the golden specification according to the task.
This creates a unified format for multimodal instruction-following sequences. GenEDA is instruction-tuned on prediction tokens using the auto-regressive training objective, maximizing the likelihood of the target ground-truth specification text sequence $y$ with length $L$, formulated as:
\begin{equation}\label{eq:ntp}
\mathcal{L}_{\text{align1}} =  - \sum_{i=1}^{L} \log p(y_i \mid y_{<i}, \mathcal{G}_\mathcal{N})
\end{equation}
where $y_{<i}$ represents the previously generated tokens before the current token $y_i$, and $\mathcal{G}_\mathcal{N}$ denotes the input netlist graph for the encoder.\looseness=-1


During instruction tuning, we leverage a two-step procedure for multimodal embedding alignment: 
\begin{enumerate}
    \item \textit{Pre-training for modality alignment.} In this stage, the netlist encoder $\mathbf{f}_\text{enc}$ and decoder $\mathbf{f}_\text{dec}$ remain frozen. Only the connector MLP $\mathbf{f}_\text{c}$ is trained to maximize the likelihood of the generated tokens of the auto-regressive loss, as formulated in~\Cref{eq:ntp}. This step aligns the netlist embeddings $H_g$ with the pre-trained LLM word embeddings $H_t$, effectively acting as a modality adapter (i.e., ``netlist tokenizer'') for the LLM.
    \item \textit{Fine-tuning end-to-end.} In this stage, the encoder $\mathbf{f}_\text{enc}$ remains frozen, while both the connector $\mathbf{f}_\text{c}$ and the LLM decoder $\mathbf{f}_\text{dec}$ are fine-tuned also with the auto-regressive loss  (i.e., \Cref{eq:ntp}). This end-to-end training step further refines the alignment, allowing the model to perform generative tasks seamlessly across graph and text modalities.
\end{enumerate}


\subsection{Paradigm 2: Prediction-Based Alignment}\label{subec:paradigm2}
\textbf{Paradigm \circled{2} overview.} In addition to trainable open-source LLMs, advanced LLMs are often frozen due to commercial or computational limitations, yet they excel in reasoning and support longer input contexts. Unlike paradigm 1, which aligns coarse-grained graph-level embeddings $N_g$, our paradigm 2 leverages fine-grained gate-level text and frozen advanced LLMs to support the more challenging exact RTL code reasoning task. As \Cref{fig:align} (b) shows, we align our multimodal encoder $\mathbf{f}_\text{enc}$ with frozen LLM decoders $\mathbf{f}_\text{dec}$ by using the encoder’s fine-grained gate functional predictions $\mathcal{T}_\text{pred}$ as textual inputs for LLMs. 
These predictions $\mathcal{T}_\text{pred}$ provide detailed gate-level analysis of netlist functionalities, which are then used by LLMs to summarize and generate high-level functionality.
This paradigm enables seamless integration without modifying the pre-trained frozen decoder. Below, we introduce the details of the task and our alignment method.\looseness=-1



\textbf{Prediction alignment architecture.}
To generate functional predictions $\mathcal{T}_\text{pred}$ with the encoder $\mathbf{f}_\text{enc}$, we fine-tune our pre-trained encoder to enable gate functionality identification with the task proposed in~\cite{alrahis2021gnn}. This task involves classifying gates into 5 high-level function types defined in their arithmetic RTL code, including adder, multiplier, comparator, subtractor, and controller.
During fine-tuning, the encoder $\mathbf{f}_\text{enc}$ remains frozen, and only the additional function predictor MLP $\mathbf{f}_\text{p}$ is trained for gate function classification using cross-entropy loss, as formulated below.\looseness=-1
\begin{equation}
\mathcal{L}_\text{align2\_pred} = - \sum_{i} y_i \log(\mathbf{f}_\text{p}(N_i)),
\end{equation}
where $y_i$ represents the ground-truth function type label for the netlist gate $i$, and $N_i$ are the encoder-generated embeddings for this gate.

After fine-tuning, the encoder generates textual predictions $\mathcal{T}_\text{pred}$ for all gates' functionalities. These predictions (i.e., gate functional roles) are directly annotated into its netlist Verilog code $\mathcal{T}_\mathcal{N}$, which serves as input text for the frozen LLM decoder. The decoder then utilizes this fine-grained gate annotation for arithmetic RTL code generation. To illustrate this process more clearly, we present a detailed case study in~\Cref{fig:task3}.
We formulate the process below:
\begin{equation}
\begin{split}
\mathbf{f}_\text{dec}(\mathcal{T}_\text{pred} + \mathcal{T}_\mathcal{N}) &\Rightarrow \text{Generative Reasoning}, \\ \text{with } \mathcal{T}_\text{pred} &= \mathbf{f}_\text{p}(\mathbf{f}_\text{enc}(\mathcal{G}_\mathcal{N})).
\end{split}
\end{equation}

\textbf{Chain-of-Thought for RTL code reasoning.}
Leveraging the annotated netlist, we employ the Chain-of-Thought (CoT) technique to improve the LLM in this challenging exact RTL code reasoning. The CoT prompt decomposes reasoning tasks into two sequential steps: (1) Reason about the word-level arithmetic function description based on the given circuit netlist with gate annotations.
(2) Generate RTL code to implement the identified arithmetic function, ensuring word-level RTL operations and avoiding bit-level operations.

\section{Experiments}\label{sec:expr}

\begin{table}[!b]
\centering
\caption{Statistics of the netlist dataset.}
\vspace{-.05in}
\resizebox{0.45\textwidth}{!}{

\begin{tabular}{c||c|c|c|c} \toprule
                                                                               & Source    & \# Circuits & \begin{tabular}[c]{@{}c@{}}\# Tokens\\      \scriptsize(avg.)\end{tabular} & \begin{tabular}[c]{@{}c@{}}\# Gates\\      \scriptsize(avg.)\end{tabular} \\ \midrule
\multirow{4}{*}{\begin{tabular}[c]{@{}c@{}}Task   \\      1 \& 2\end{tabular}} & ITC99~\cite{corno2000rt}     & 4k          & 15k                                                             & 1025                                                           \\
                                                                               & OpenCores~\cite{URL:opencore} & 55k         & 9k                                                              & 173                                                            \\
                                                                               & Chipyard~\cite{amid2020chipyard}  & 20k         & 24k                                                             & 2813                                                           \\
                                                                               & VexRiscv~\cite{vexriscv}  & 21k         & 13k                                                             & 901                                                            \\ \midrule
Task 3                                                                         & GNN-RE~\cite{alrahis2021gnn}    & 8           & 4k                                                              & 67       \\ \bottomrule                                                     
\end{tabular}

}

\label{tbl:stat}
\vspace{-.1in}
\end{table}

\begin{table*}[!t]
\vspace{-.2in}
\center
\caption{Evaluation results on Task 1 \& 2, reasoning specification text from gate-level netlists. Best results are highlighted in bold.}
\vspace{-.05in}
\resizebox{1\textwidth}{!}{

\begin{tabular}{ccccccc||cccccc} \toprule
\multicolumn{1}{l}{}   & \multicolumn{6}{c||}{\textbf{Task 1 Functional   Description Reasoning}}                         & \multicolumn{6}{c}{\textbf{Task 2 Implementation Detail Reasoning}}                            \\ \cmidrule{2-13}
\textbf{Model }                 & BLEU        & ROUGE-1     & ROUGE-2     & ROUGE-L     & Emb. Sim.     & GPT Score     & BLEU        & ROUGE-1     & ROUGE-2     & ROUGE-L     & Emb. Sim.     & GPT Score     \\ \midrule
GPT4o                  & 5           & 34          & 10          & 17          & 0.83          & 0.21          & 5           & 34          & 10          & 17          & 0.84          & 0.44          \\
DeepSeek-V3     & 4           & 31          & 9           & 17          & 0.82          & 0.19          & 3           & 34          & 10          & 17          & 0.83          & 0.42          \\ \midrule
DeepSeek-1B      & 0           & 7           & 1           & 5           & 0.77          & 0.04          & 0           & 5           & 1           & 4           & 0.73          & 0.01          \\
DeepSeek-7B      & 0           & 8           & 2           & 6           & 0.77          & 0.06          & 0           & 5           & 1           & 4           & 0.74          & 0.02          \\ \midrule
RTLCoder-7B$^\star$      & 0           & 5           & 1           & 4           & 0.32          & 0.02          & 0           & 3           & 1           & 2           & 0.28          & 0.01          \\ \midrule
\cellcolor[HTML]{e0ebf6}\textbf{GenEDA-1B} & 12 & 47 & 16 & 27 & 0.9  & 0.62 & 12 & 47 & 19 & 28 & 0.93 & 0.5  \\
\cellcolor[HTML]{e0ebf6}\textbf{GenEDA-7B} & \textbf{14} & \textbf{49} & \textbf{19} & \textbf{28} & \textbf{0.91} & \textbf{0.62} & \textbf{14} & \textbf{50} & \textbf{21} & \textbf{30} & \textbf{0.94} & \textbf{0.51} \\ \bottomrule
\end{tabular}

}
\begin{tablenotes}\footnotesize
\item $^\star$ RTLCoder~\cite{liu2024rtlcoder} is fine-tuned for generating RTL code from specifications. In our setting, we adopt the fine-tuned model as a circuit-specific LLM baseline to generate specifications from netlists reversely.
\end{tablenotes} 
\label{tbl:task12}
\end{table*}

\begin{figure*}[!t]
  \centering
  \includegraphics[width=1\linewidth]{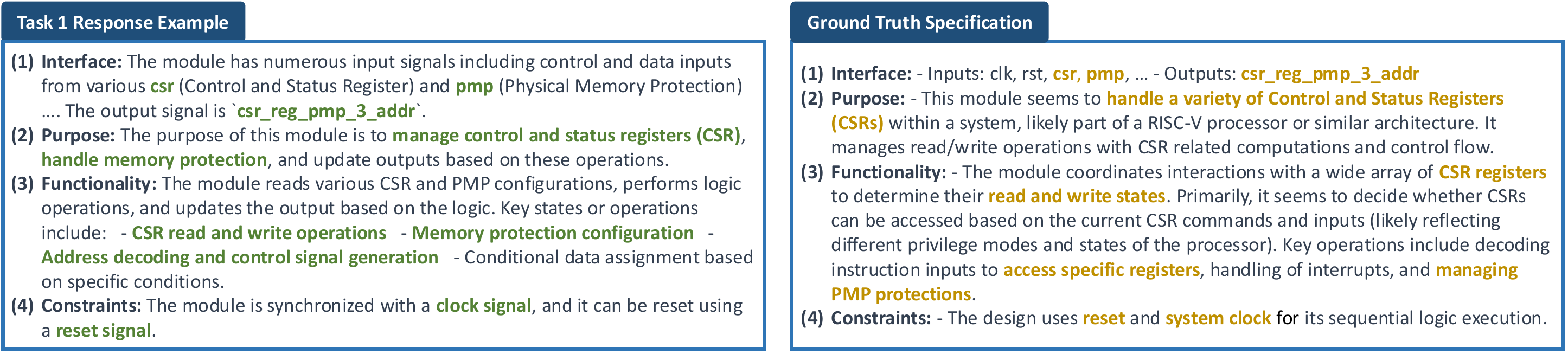}
    \vspace{-.15in}
  \caption{Case study for Task 1. Comparison between the response from GenEDA and the ground truth specification for a circuit module.} 
  \label{fig:task1}
  \vspace{-.05in}
\end{figure*}


\begin{figure}[!b]
  \centering
  \includegraphics[width=1\linewidth]{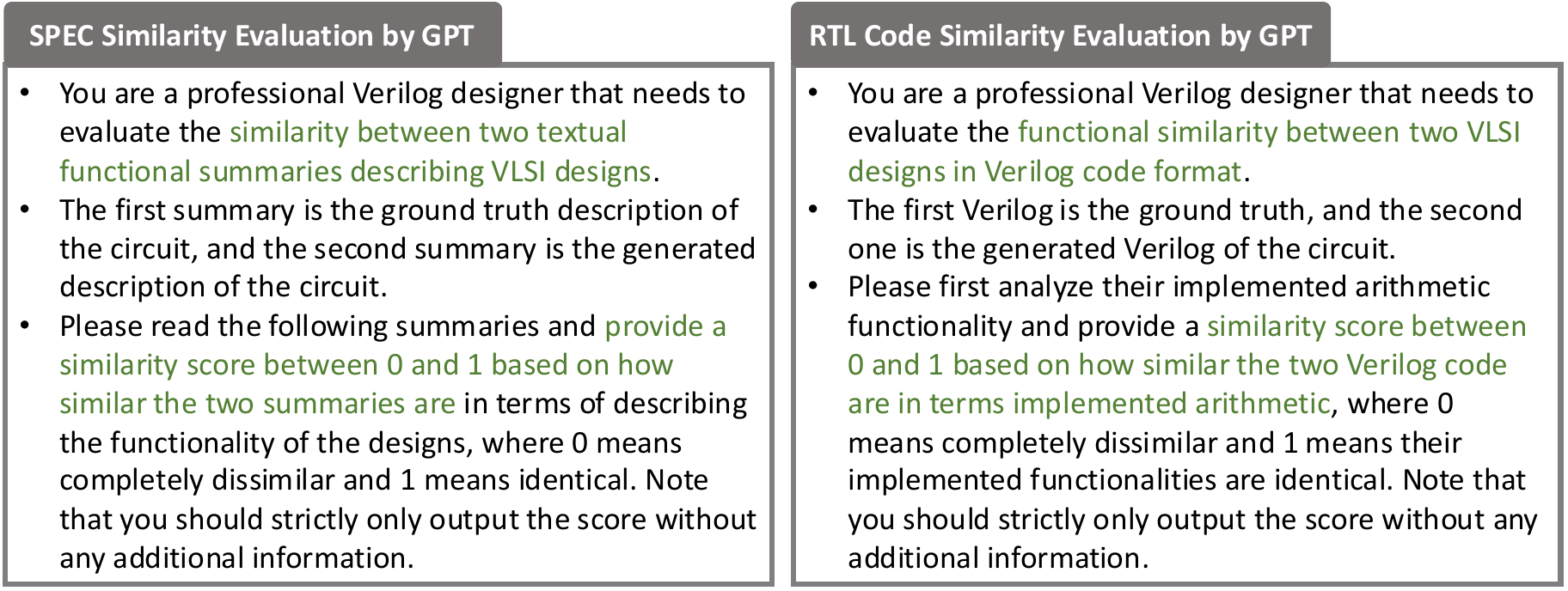}
  \caption{Prompt used in GPT-assisted evaluation (i.e., GPT Score).} 
  \label{fig:eval}
\end{figure}

In this section, we first introduce the experimental setup and evaluation metrics in~\Cref{expr:setup}. Then, in~\Cref{expr:task1}, we present our results on specification context reasoning (i.e., Task 1 \& 2). Then we discuss the arithmetic RTL code reasoning results (i.e., Task 3) in~\Cref{expr:rtl} and conclude with the ablation study in~\Cref{expr:abl}.

\subsection{Experimental Settings}
\label{expr:setup}

\textbf{Circuit dataset preparation.}
For the circuit encoder and Task 1 \& 2, we collect circuit datasets from various open-source RTL code benchmarks, including ITC99~\cite{corno2000rt}, OpenCores~\cite{URL:opencore}, Chipyard~\cite{amid2020chipyard}, and VexRiscv~\cite{vexriscv}. In Task 1 and 2, for large sequential circuits, we split them into multiple sub-circuits, following the techniques in~\cite{fang2025nettag}. After splitting, we collect 25k subcircuits, which are augmented by functional equivalent transformation by Yosys~\cite{wolf2013yosys} to create a total of 50k samples. We randomly sample 400 subcircuits for testing, ensuring that no subcircuits from the same circuit are included in the training set.
Since no specification documents are available, we generate functional specifications from the RTL context using GPT-4o as the ground-truth for instruction-tuning. Human engineers then verify these generated specifications, and those that do not contain valid functionality are excluded.
For Task 3, we use the open-source arithmetic RTL code from~\cite{alrahis2021gnn}. Please note that in all tasks, only low-level netlists are provided for reverse reasoning, with high-level specifications and RTL code being unknown to models.
All RTL designs are synthesized into netlists using the Synopsys Design Compiler with the NanGate 45nm technology library. 
We provide detailed statistics of our netlist dataset in~\Cref{tbl:stat}, including the number of circuits, the average number of text tokens, and the average number of gates.

\begin{table*}[!t]
\center
\vspace{-.16in}
\caption{Evaluation results on Task 3, reasoning arithmetic RTL code from gate-level netlists. Each design is generated five times per model. Best results are highlighted in bold.}
\vspace{-.05in}
\resizebox{1\textwidth}{!}{

\begin{tabular}{c|ccc|ccc||ccc|ccc} \toprule
                      & \multicolumn{3}{c|}{\textbf{GPT-4o}} & \multicolumn{3}{c||}{\cellcolor[HTML]{e0ebf6}\textbf{GenEDA (w. GPT-4o)}} & \multicolumn{3}{c|}{\textbf{DeepSeek-V3}} & \multicolumn{3}{c}{\cellcolor[HTML]{e0ebf6}\textbf{GenEDA (w.   DeepSeek-V3)}} \\ \cmidrule{2-13}
\textbf{Circuit}      & Syntax& GPT Score   & Function& Syntax& GPT Score   & Function& Syntax& GPT Score   & Function& Syntax& GPT Score   & Function\\ \midrule
1                     & 80\%
& 0.36        & 0\%          & 100\%              & 0.52             & 20\%               & 100\%          & 0.18          & 0\%           & 100\%                & 0.85                & 80\%                 \\
2                     & 20\%
& 0.32        & 0\%          & 100\%              & 0.88             & 80\%               & 100\%          & 0.55          & 40\%           & 100\%                & 0.99                & 100\%                 \\
3                     & 100\%
& 0.3         & 60\%          & 100\%              & 0.28             & 40\%               & 100\%          & 0.2           & 0\%           & 100\%                & 0.74                & 60\%                 \\
4                     & 100\%
& 0.66        & 60\%          & 40\%              & 0.44             & 20\%               & 100\%          & 0.95          & 100\%           & 100\%                & 1                   & 100\%                 \\
5                     & 60\%
& 0.18        & 0\%          & 80\%             & 0.51             & 0\%               & 100\%          & 0             & 0\%           & 100\%                & 0.98                & 80\%                 \\
6                     & 20\%
& 0.2         & 0\%          & 100\%              & 0.79             & 60\%               & 100\%          & 0.28          & 0\%           & 100\%               & 0.95                & 100\%                 \\
7                     & 80\%
& 0.18        & 0\%          & 80\%              & 0.54             & 0\%               & 60\%          & 0.18          & 0\%           & 80\%                & 0.91                & 0\%                 \\
8                     & 40\%
& 0.28        & 0\%          & 80\%              & 0.7              &0\%               & 40\%          & 0.49          & 0\%           & 80\%                & 0.76                & 0\%                 \\
9                     & 80\%
& 0.3         & 0\%          & 100\%              & 0.56             &40\%               & 100\%          & 0.2           & 0\%           & 100\%                & 0.78                & 100\%                 \\ \midrule
\textbf{Success Rate} & 64\%& 0.31        & 13\%& 87\%& 0.58& 29\%& 89\%& 0.34          & 16\%& \textbf{96\%}& \textbf{0.88}                & \textbf{69\%}\\
\textbf{Pass@1}       & 67\%         & /            &  17\%          &  83\%          &  /             &  50\%           &   92\%      &/       &  8\%                &  \textbf{100\%}               &   /                                   &  \textbf{58\%}                 \\
\textbf{Pass@5}       & 100\%         &    /         &  22\%          & 100\%           &     /          &   67\%          &    100\%            &   /               &  22\%               & \textbf{100\%}                 & /                    & \textbf{78\%}             \\ \bottomrule     
\end{tabular}

}
\label{tbl:task3}
\end{table*}

\begin{figure*}[!t]
  \centering
  \includegraphics[width=1\linewidth]{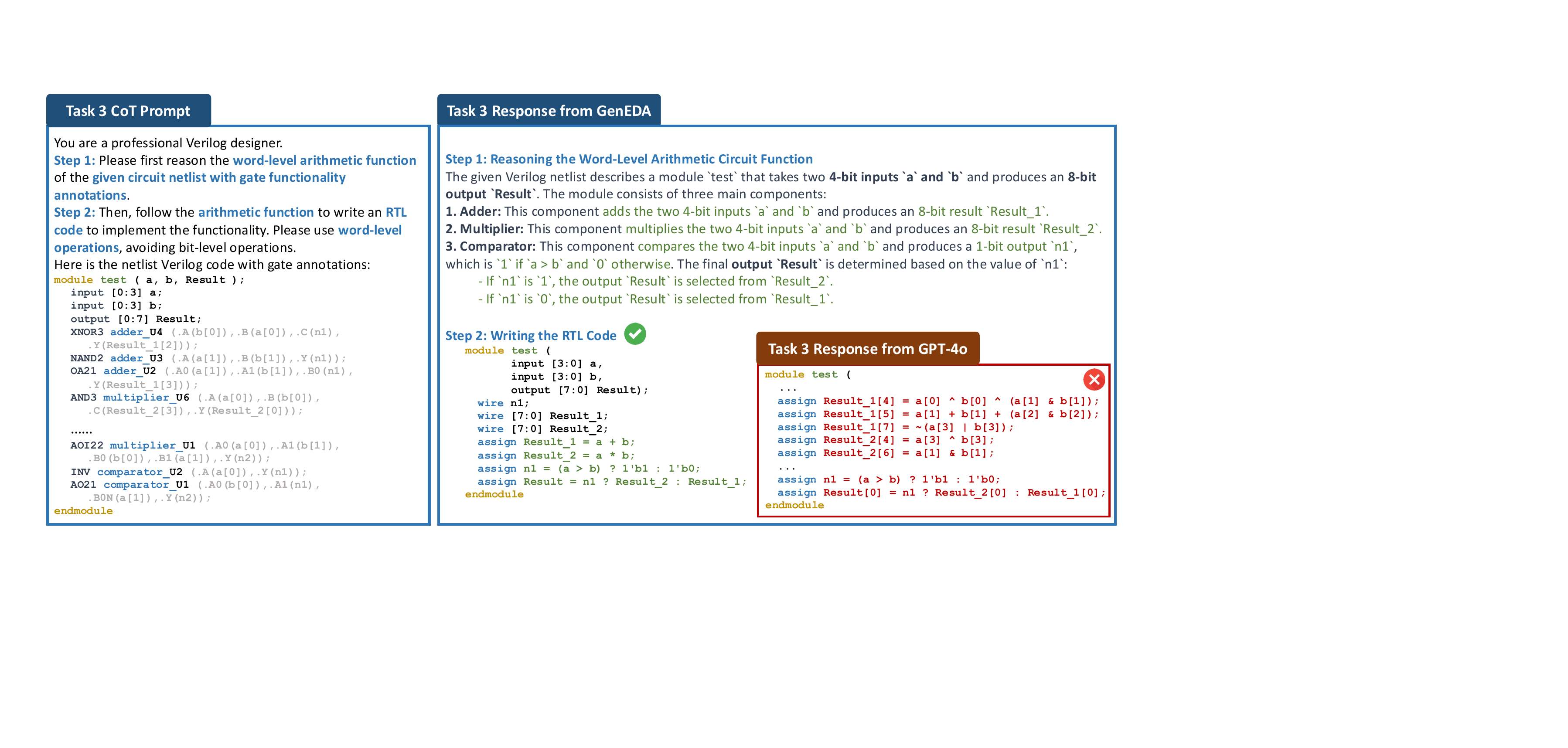}
  \vspace{-.2in}
  \caption{Case study for Task 3. By utilizing gate function predictions from GenEDA’s encoder, the LLM in GenEDA can reason about the \textit{word-level} arithmetic functionality of bit-blasted netlists and accurately reconstruct the corresponding RTL code (blue box). In contrast, without these predictions, LLMs (e.g., GPT-4o) generate only \textit{bit-level} operations (red box), leading to significantly lower reconstruction accuracy.\looseness=-1}
  \label{fig:task3}
  \vspace{-.1in}
\end{figure*}

\textbf{Model and training.} For the encoder model in GenEDA, we employ the state-of-the-art netlist encoder NetTAG~\cite{fang2025nettag} as the backbone. For the decoder LLMs in GenEDA, we fine-tune the DeepSeek-Coder~\cite{guo2024deepseek} 1B and 7B models in the trainable embedding-based alignment (Paradigm 1), and directly leverage the commercial APIs of OpenAI GPT-4o and DeepSeek-V3~\cite{liu2024deepseek} as the frozen LLMs in the prediction-based alignment (Paradigm 2). The training process utilizes DeepSpeed ZeRO and LoRA techniques.
In alignment paradigm 1, we adopt a three-layer MLP with dimensions 768, 2048, and 4096 to transform the netlist embedding (768 dimensions) into the LLM word embeddings (4096 dimensions). This connector can be further explored using more advanced methods, such as Q-Former and cross-attention mechanism as in~\cite{liu2024visual, li2023blip, li2022blip}\footnote{Complex connectors like Q-Former are typically used with frozen open-source LLMs~\cite{li2023blip, li2022blip}, while simpler connectors such as MLPs are often suitable when the LLM is trainable~\cite{liu2024visual}.}. GenEDA is instruction-tuned using LoRA on the full task-specific dataset for one epoch.
In alignment paradigm 2, the function predictor MLP for encoder fine-tuning contains three layers with a hidden dimension of 256.
Experiments are conducted on a server equipped with 8 NVIDIA A800 80G PCIe GPUs.\looseness=-1

\subsection{Benchmark Evaluation}

For specification reasoning tasks (Task 1 \& 2), the model-generated natural language specification is compared directly with the ground-truth specification using both natural language similarity metrics and LLM-assisted functionality evaluation metrics. Specifically, natural language similarity scores, including BLEU, ROUGE-1/2/L, are computed.
These metrics assess the overlap between the generated text and the reference text by comparing n-grams.
For LLM-based evaluation, we use the OpenAI text embedding model, text-embedding-ada-002, to obtain text embeddings for both the generated and reference specifications. Cosine similarity is then calculated between the embeddings to generate the embedding similarity score. Additionally, we utilize GPT-4o to directly evaluate the specifications and assign a similarity score between 0 and 1 based on how closely the generated specification matches the intended functionality of the designs. Detailed prompts for GPT-4o are provided in~\Cref{fig:eval}.

For RTL code reasoning (Task 3), we use Synopsys VCS to simulate the generated RTL code with the proposed testbench, validating both syntax and functional correctness. Each circuit is generated five times, and we compute the average success rate and Pass@k metrics~\cite{liu2024rtlcoder}. Similar to specification reasoning, GPT-4o is also used to evaluate the functional similarity between the generated and ground-truth RTL codes, with the prompt outlined in~\Cref{fig:eval}.

\subsection{Result of Specification Reasoning (Task 1 \& 2)}

\textbf{Baseline solutions.} To ensure a comprehensive comparison, we evaluate both general-purpose and circuit-adapted LLMs as baseline methods. For general-purpose LLMs, we use advanced commercial models like GPT-4o and DeepSeek-V3, as well as lightweight open-source models like DeepSeek-Coder 1B and 7B. For circuit-adapted LLMs, we select RTLCoder-7B~\cite{liu2024rtlcoder},  a representative LLM fine-tuned for spec-to-RTL generation. These baselines take netlist text and the question prompt as input and generate the specification text.

\textbf{Comparison with baselines.}
\label{expr:task1}
\Cref{tbl:task12} presents the evaluation results for specification reasoning in Task 1 and Task 2, comparing GenEDA models with baseline models. 
For both tasks, after aligning with netlist encoder embeddings through multimodal instruction tuning, our GenEDA models (1B and 7B), based on DeepSeek Coder 1B and 7B, significantly outperform all general-purpose LLMs across all textual semantic similarity metrics. 
Additionally, although RTLCoder~\cite{liu2024rtlcoder} is fine-tuned for RTL code generation from specifications, it performs poorly on both tasks, even underperforming its base model, DeepSeek-7B. This is primarily due to the substantial differences between the tasks.
Notably, for the GPT scores, which analyze the similarity between generated specifications and ground truth with GPT, GenEDA scored much higher than the baseline LLMs.\looseness=-1 

These results highlight the effectiveness of aligning circuit structural and functional information through encoders to enhance generative capabilities.
Moreover, the GenEDA-7B demonstrates further improvements over the GenEDA-1B, indicating that potential gains can be achieved by employing more powerful open-sourced base models.\looseness=-1

\textbf{Natural language specification reasoning case study.}
We present detailed case studies for these results in~\Cref{fig:task1}.
We compare a model-generated response and the corresponding ground truth specification for Task 1, function description generation. The model accurately generates key functionality of the specification, aligning closely with the ground truth. For example, in the functionality section, the model effectively describes how the module handles various control and status registers and memory protection configuration, which matches the ground truth’s detailed explanation of register states and operations. These results underline GenEDA’s capability to generate high-level natural language descriptions from low-level netlist inputs.

\subsection{Result of RTL Code Reasoning (Task 3)}
\label{expr:rtl}

\textbf{Baseline solutions.} In this task, we choose the advanced commercial LLMs, including GPT-4o and DeepSeek-V3, as the baseline methods. For the open-source LLMs (e.g., DeepSeek-Coder 1B, 7B, and RTLCoder), these models fail to generate high-level RTL and instead produce only gate-level netlist code.



\textbf{Comparison with baselines.}
For netlist gate function classification, our encoder achieves a 97\% accuracy rate, providing a strong foundation for our prediction-based alignment paradigm. The impact of encoder quality on alignment performance is further discussed in~\Cref{expr:abl}.
\Cref{tbl:task3} evaluates Task 3: Arithmetic RTL Code Reasoning for various models. GenEDA combined with DeepSeek-V3 achieves the highest success rate with a 97\% syntax pass rate, 88\% GPT Score, and a 60\% functional pass rate, outperforming GPT-4o, DeepSeek-V3 alone, and other combinations. Without the prediction alignment, the powerful commercial LLMs alone cannot reason the high-level word-level arithmetics to generate RTL code, and they only generate RTL code using bit-level operations.
This demonstrates GenEDA’s ability to reversely reconstruct RTL code from netlists with exact functionality. 

Notably, even state-of-the-art commercial LLMs like GPT-4o and DeepSeek-V3 achieve less than a 20\% functional success rate on our benchmark, highlighting the difficulty of this task. In contrast, existing specification-to-RTL generation benchmarks like RTLLM~\cite{lu2024rtllm} and VerilogEval~\cite{liu2023verilogeval} report over 60\% success with off-the-shelf LLMs like GPT-4, emphasizing the challenge of reverse reconstructing high-level RTL code from low-level bit-blasted netlists.


\textbf{Arithmetic RTL code reasoning case study.}
\Cref{fig:task3} provides an example of the reasoning process for Task 3. The Chain-of-Thought (CoT) prompt guides the model in two steps: (1) Understanding the arithmetic circuit function: The model reasons about the circuit’s gate annotations to identify components like adders, multipliers, and comparators, and determines their combined functionality. These predictions are then annotated back onto the original netlist text, which is provided to the LLM as input for further reasoning.
(2) Writing the RTL code: Based on the identified functionality, the model generates RTL code using word-level operations, successfully implementing the circuit’s logic.
This case study illustrates the effectiveness of GenEDA in generating correct and interpretable RTL code, bridging low-level gate details with high-level functional implementations.\looseness=-1


\begin{figure}[!t]
  \centering
  \includegraphics[width=1\linewidth]{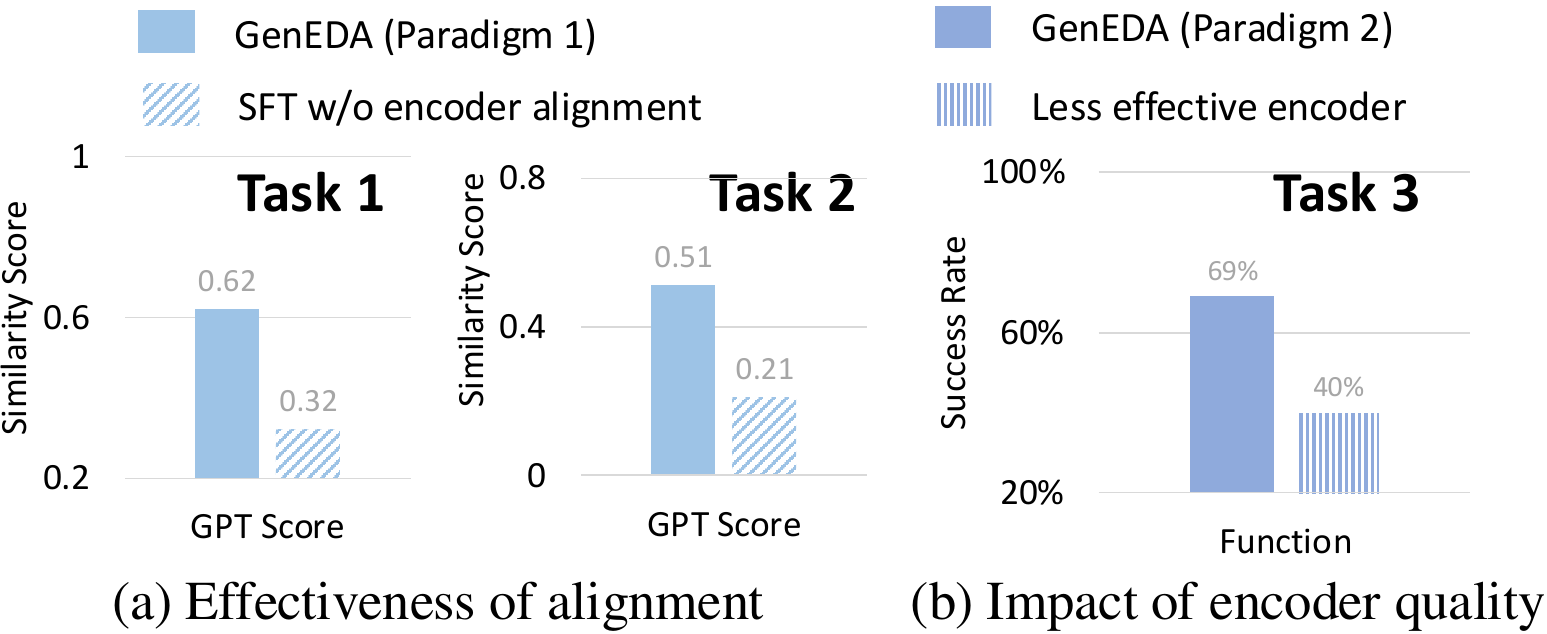}
  \vspace{-.15in}
  \caption{Ablation study demonstrating the effectiveness of encoder-decoder alignment and the impact of encoder quality.} 
  \label{fig:abl}
  \vspace{-.15in}
\end{figure}

\subsection{Ablation Study}
\label{expr:abl}
In~\Cref{fig:abl}, we present ablation studies to demonstrate the effectiveness of our encoder-decoder alignment and encoder quality in improving generative reasoning tasks. We conduct experiments by selectively removing the encoder alignment or using less effective encoders to assess their impact on task performance. These studies allow us to isolate and understand the contributions of different components of GenEDA to its overall performance.

\textbf{Effectiveness of alignment with encoders.} 
GenEDA alignment paradigm 1 is achieved by cross-modally fine-tuning the LLMs using netlist encoder embeddings. In this ablation study, we remove the encoder alignment and only perform supervised fine-tuning (SFT) of the LLMs using the same prompts and labels as in GenEDA. As shown in~\Cref{fig:abl} (a), removing the encoder alignment significantly decreases model performance on Task 1 and Task 2 across all metrics. Notably, the GPT scores drop sharply from 0.62 to 0.32 on Task 1 and from 0.51 to 0.21 on Task 2, highlighting the effectiveness of embedding alignment.\looseness=-1

\textbf{Impact of encoder quality.}
GenEDA alignment paradigm 2 heavily relies on the accuracy of gate functionality classification from the encoder. In this ablation study, we replace the high-quality encoder NetTAG~\cite{fang2025nettag} with a less effective baseline, GNN-RE~\cite{alrahis2021gnn}. This results in a significant drop in classification accuracy from 97\% to 83\%. Consequently, the performance of reconstructed RTL code also degrades, with syntax accuracy dropping from 96\% to 91\% and function accuracy decreasing from 69\% to only 40\%, as shown in~\Cref{fig:abl} (b). This demonstrates the importance of using high-quality encoders and their impact on the generation after alignment.

\section{Discussion} \label{sec:discussion}

\subsection{Extending GenEDA Alignment to Other Circuit Stages and Tasks}
Beyond the netlist stage addressed in this work, GenEDA’s encoder-decoder alignment framework can be extended to various stages in the circuit design flow. At the RTL stage, even though the same RTL functionality can be generated, different circuit structures can yield significantly varying PPA characteristics. By leveraging structural RTL information captured by RTL encoders, LLMs can potentially enable structure-aware generation, generating more optimized RTL code with better PPA metrics.
Additionally, at the layout stage, GenEDA can be adapted to handle cross-modal inputs, such as layout images and netlist graphs. This might enable the direct generation of macro positions on a chip by learning from image representations, thus enabling more efficient and optimized physical design generation.

\subsection{Potential Application of Generative Netlist Functional Reasoning}
GenEDA can reason the detailed functionality from netlists, which can significantly benefit verification and optimization processes.
Before loading the netlists into EDA tools, GenEDA can guide the selection of appropriate strategies for verifying or optimizing different parts of the design, such as datapaths or control logic. Additionally, it can assist in verifying the equivalence of netlists by transforming them into higher-level RTL code, making the verification process more scalable.
In hardware security, GenEDA can be applied to detect malicious hardware trojans. By analyzing netlists, it can identify unexpected or unauthorized functional behaviors, helping to ensure the integrity and security of hardware systems.

\section{Conclusion and Future Work}

In this paper, we present GenEDA, a framework that aligns multimodal circuit encoders and decoders for advanced generative netlist functional reasoning tasks. We align the state-of-the-art netlist encoder with both trainable and frozen LLM decoders through two alignment paradigms. Our experiments and ablation studies demonstrate the effectiveness of this approach, with GenEDA significantly enhancing the performance of state-of-the-art LLMs, showcasing the critical role of integrating both graph and text circuit modalities for complex netlist tasks.
Future work will explore extending this alignment framework to other stages of the circuit design flow, such as RTL code generation and layout-stage tasks, further enhancing the capabilities of GenEDA for diverse EDA applications.


\section{Acknowledgement}
This research was supported by Hong Kong Research Grants Council (RGC) CRF-YCRG Grant C6003-24Y, GRF 16200724, and ACCESS – AI Chip Center for Emerging Smart Systems, sponsored by the InnoHK initiative of the Innovation and Technology Commission of the Hong Kong Special Administrative Region Government.

\clearpage
\bibliographystyle{IEEEtran}
\bibliography{ref}

\begin{thebibliography}{10}
\providecommand{\url}[1]{#1}
\csname url@samestyle\endcsname
\providecommand{\newblock}{\relax}
\providecommand{\bibinfo}[2]{#2}
\providecommand{\BIBentrySTDinterwordspacing}{\spaceskip=0pt\relax}
\providecommand{\BIBentryALTinterwordstretchfactor}{4}
\providecommand{\BIBentryALTinterwordspacing}{\spaceskip=\fontdimen2\font plus
\BIBentryALTinterwordstretchfactor\fontdimen3\font minus \fontdimen4\font\relax}
\providecommand{\BIBforeignlanguage}[2]{{%
\expandafter\ifx\csname l@#1\endcsname\relax
\typeout{** WARNING: IEEEtran.bst: No hyphenation pattern has been}%
\typeout{** loaded for the language `#1'. Using the pattern for}%
\typeout{** the default language instead.}%
\else
\language=\csname l@#1\endcsname
\fi
#2}}
\providecommand{\BIBdecl}{\relax}
\BIBdecl

\bibitem{chen2024large}
L.~Chen, Y.~Chen, Z.~Chu \emph{et~al.}, ``Large circuit models: opportunities and challenges,'' \emph{Science China Information Sciences}, 2024.

\bibitem{fang2025cfm}
W.~Fang, J.~Wang, Y.~Lu, S.~Liu, Y.~Wu, Y.~Ma, and Z.~Xie, ``A survey of circuit foundation model: Foundation ai models for vlsi circuit design and eda,'' \emph{arXiv preprint arXiv:2504.03711}, 2025.

\bibitem{fang2025circuitencoder}
W.~Fang, S.~Liu, H.~Zhang, and Z.~Xie, ``A self-supervised, pre-trained, and cross-stage-aligned circuit encoder provides a foundation for various design tasks,'' in \emph{ASP-DAC}, 2025.

\bibitem{shi2024deepgate3}
Z.~Shi, Z.~Zheng, S.~Khan, J.~Zhong, M.~Li, and Q.~Xu, ``Deepgate3: towards scalable circuit representation learning,'' \emph{arXiv preprint arXiv:2407.11095}, 2024.

\bibitem{shi2023deepgate2}
Z.~Shi, H.~Pan, S.~Khan, M.~Li, Y.~Liu, J.~Huang, H.-L. Zhen, M.~Yuan, Z.~Chu, and Q.~Xu, ``{DeepGate2}: Functionality-aware circuit representation learning,'' in \emph{ICCAD}, 2023.

\bibitem{li2022deepgate}
M.~Li, S.~Khan, Z.~Shi, N.~Wang, H.~Yu, and Q.~Xu, ``{DeepGate}: Learning neural representations of logic gates,'' in \emph{DAC}, 2022.

\bibitem{wang2022functionality}
Z.~Wang, C.~Bai, Z.~He, G.~Zhang, Q.~Xu, T.-Y. Ho, B.~Yu, and Y.~Huang, ``Functionality matters in netlist representation learning,'' in \emph{Design Automation Conference (DAC)}, 2022.

\bibitem{fang2025circuitfusion}
W.~Fang, S.~Liu, J.~Wang, and Z.~Xie, ``Circuitfusion: multimodal circuit representation learning for agile chip design,'' in \emph{International Conference on Learning Representations (ICLR)}, 2025.

\bibitem{wu2025circuit}
H.~Wu, H.~Zheng, Y.~Pu, and B.~Yu, ``Circuit representationlearning with masked gatemodeling and verilog-aigalignment,'' in \emph{International Conference on Learning Representations (ICLR)}, 2025.

\bibitem{deng2024less}
C.~Deng, Z.~Yue \emph{et~al.}, ``Less is more: Hop-wise graph attention for scalable and generalizable learning on circuits,'' in \emph{DAC}, 2024.

\bibitem{xu2023fast}
C.~Xu, P.~Sharma, T.~Wang, and L.~W. Wills, ``Fast, robust and transferable prediction for hardware logic synthesis,'' in \emph{IEEE/ACM International Symposium on Microarchitecture (MICRO)}, 2023.

\bibitem{zhang2024mg}
Y.~Zhang, Z.~Yu \emph{et~al.}, ``Mg-verilog: Multi-grained dataset towards enhanced llm-assisted verilog generation,'' \emph{arXiv preprint arXiv:2407.01910}, 2024.

\bibitem{pei2024betterv}
Z.~Pei, H.-L. Zhen, M.~Yuan, Y.~Huang, and B.~Yu, ``Betterv: Controlled verilog generation with discriminative guidance,'' \emph{arXiv preprint arXiv:2402.03375}, 2024.

\bibitem{liu2023rtlcoder}
S.~Liu, W.~Fang, Y.~Lu, Q.~Zhang, H.~Zhang, and Z.~Xie, ``{RTLCoder: Outperforming GPT-3.5 in design RTL generation with our open-source dataset and lightweight solution},'' \emph{IEEE LLM-Aided Design (LAD)}, 2023.

\bibitem{fang2024assertllm}
Z.~Yan, W.~Fang, M.~Li, M.~Li, Z.~Yan, S.~Liu, Z.~Xie, and H.~Zhang, ``{AssertLLM}: Generating and evaluating hardware verification assertions from design specifications via multi-{LLM}s,'' in \emph{ASP-DAC}, 2025.

\bibitem{kande2024security}
R.~Kande, H.~Pearce \emph{et~al.}, ``(security) assertions by large language models,'' \emph{IEEE Transactions on Information Forensics and Security (TIFS)}, 2024.

\bibitem{he2023chateda}
Z.~He, H.~Wu, X.~Zhang, X.~Yao, S.~Zheng, H.~Zheng, and B.~Yu, ``Chateda: A large language model powered autonomous agent for eda,'' in \emph{Workshop on Machine Learning for CAD (MLCAD)}, 2023.

\bibitem{zheng2025deepgate4}
Z.~Zheng, S.~Huang, J.~Zhong \emph{et~al.}, ``Deepgate4: Efficient and effective representation learning for circuit design at scale,'' in \emph{International Conference on Learning Representations (ICLR)}, 2025.

\bibitem{fang2025nettag}
W.~Fang, W.~Li, S.~Liu, Y.~Lu, H.~Zhang, and Z.~Xie, ``Nettag: A multimodal rtl-and-layout-aligned netlist foundation model via text-attributed graph,'' in \emph{Design Automation Conference (DAC)}, 2025.

\bibitem{liu2024rtlcoder}
S.~Liu, W.~Fang, Y.~Lu, Q.~Zhang, H.~Zhang, and Z.~Xie, ``Rtlcoder: Fully open-source and efficient llm-assisted rtl code generation technique,'' \emph{IEEE TCAD}, 2024.

\bibitem{yao2024hdldebugger}
X.~Yao, H.~Li, T.~H. Chan, W.~Xiao, M.~Yuan, Y.~Huang, L.~Chen, and B.~Yu, ``Hdldebugger: Streamlining hdl debugging with large language models,'' \emph{arXiv preprint arXiv:2403.11671}, 2024.

\bibitem{liu2025deeprtl}
Y.~Liu, C.~Xu, Y.~Zhou, Z.~Li, and Q.~Xu, ``Deep{RTL}: Bridging verilog understanding and generation with a unified representation model,'' in \emph{International Conference on Learning Representations (ICLR)}, 2025.

\bibitem{wu2023gamora}
N.~Wu, Y.~Li, C.~Hao, S.~Dai, C.~Yu, and Y.~Xie, ``Gamora: Graph learning based symbolic reasoning for large-scale boolean networks,'' in \emph{ACM/IEEE Design Automation Conference (DAC)}, 2023.

\bibitem{wang2024fgnn2}
Z.~Wang, C.~Bai, Z.~He, G.~Zhang, Q.~Xu, T.-Y. Ho, Y.~Huang, and B.~Yu, ``Fgnn2: A powerful pre-training framework for learning the logic functionality of circuits,'' \emph{IEEE Transactions on Computer-Aided Design of Integrated Circuits and Systems (TCAD)}, 2024.

\bibitem{kuehlmann2002robust}
A.~Kuehlmann, V.~Paruthi, F.~Krohm, and M.~K. Ganai, ``Robust boolean reasoning for equivalence checking and functional property verification,'' \emph{IEEE TCAD}, 2002.

\bibitem{subramanyan2013reverse}
P.~Subramanyan, N.~Tsiskaridze, K.~Pasricha, D.~Reisman, A.~Susnea, and S.~Malik, ``Reverse engineering digital circuits using functional analysis,'' in \emph{DATE}, 2013.

\bibitem{khan2025deepseq2}
S.~Khan, Z.~Shi, Z.~Zheng, M.~Li, and Q.~Xu, ``Deepseq2: Enhanced sequential circuit learning with disentangled representations,'' in \emph{Asia and South Pacific Design Automation Conference (ASP-DAC)}, 2025.

\bibitem{PolarGate}
J.~Liu, J.~Zhai, M.~Zhao, Z.~Lin, B.~Yu, and C.~Shi, ``Polargate: Breaking the functionality representation bottleneck of and-inverter graph neural network,'' in \emph{ICCAD}, 2024.

\bibitem{liu2023chipnemo}
M.~Liu, T.-D. Ene, R.~Kirby, C.~Cheng, N.~Pinckney \emph{et~al.}, ``{ChipNeMo: Domain-Adapted LLMs for Chip Design},'' \emph{arXiv preprint arXiv:2311.00176}, 2023.

\bibitem{chang2023chipgpt}
K.~Chang, Y.~Wang, H.~Ren, M.~Wang, S.~Liang, Y.~Han, H.~Li, and X.~Li, ``Chipgpt: How far are we from natural language hardware design,'' \emph{arXiv preprint arXiv:2305.14019}, 2023.

\bibitem{sun2023towards}
C.~Sun, C.~Hahn, and C.~Trippel, ``Towards improving verification productivity with circuit-aware translation of natural language to systemverilog assertions,'' in \emph{International Workshop on Deep Learning-aided Verification (DAV)}, 2023.

\bibitem{yao2024rtlrewriter}
X.~Yao, Y.~Wang, X.~Li, Y.~Lian, R.~Chen, L.~Chen, M.~Yuan, H.~Xu, and B.~Yu, ``Rtlrewriter: Methodologies for large models aided rtl code optimization,'' \emph{arXiv preprint arXiv:2409.11414}, 2024.

\bibitem{jiang2024fabgpt}
Y.~Jiang, X.~Lu, Q.~Jin, Q.~Sun, H.~Wu, and C.~Zhuo, ``Fabgpt: An efficient large multimodal model for complex wafer defect knowledge queries,'' \emph{arXiv preprint arXiv:2407.10810}, 2024.

\bibitem{wu2024chateda}
H.~Wu, Z.~He \emph{et~al.}, ``Chateda: A large language model powered autonomous agent for eda,'' \emph{IEEE Transactions on Computer-Aided Design of Integrated Circuits and Systems}, 2024.

\bibitem{li2012reverse}
W.~Li, Z.~Wasson, and S.~A. Seshia, ``Reverse engineering circuits using behavioral pattern mining,'' in \emph{2012 IEEE international symposium on hardware-oriented security and trust}.\hskip 1em plus 0.5em minus 0.4em\relax IEEE, 2012, pp. 83--88.

\bibitem{gascon2014template}
A.~Gasc{\'o}n, P.~Subramanyan, B.~Dutertre, A.~Tiwari, D.~Jovanovi{\'c}, and S.~Malik, ``Template-based circuit understanding,'' in \emph{Formal Methods in Computer-Aided Design (FMCAD)}.\hskip 1em plus 0.5em minus 0.4em\relax IEEE, 2014, pp. 83--90.

\bibitem{alrahis2021gnn}
L.~Alrahis, A.~Sengupta \emph{et~al.}, ``Gnn-re: Graph neural networks for reverse engineering of gate-level netlists,'' \emph{IEEE TCAD}, 2021.

\bibitem{chowdhury2021reignn}
S.~D. Chowdhury, K.~Yang, and P.~Nuzzo, ``Reignn: State register identification using graph neural networks for circuit reverse engineering,'' in \emph{ICCAD}, 2021.

\bibitem{lu2024rtllm}
Y.~Lu, S.~Liu, Q.~Zhang, and Z.~Xie, ``{RTLLM}: An open-source benchmark for design rtl generation with large language model,'' in \emph{Asia and South Pacific Design Automation Conference (ASP-DAC)}, 2024.

\bibitem{liu2023verilogeval}
M.~Liu, N.~Pinckney, B.~Khailany, and H.~Ren, ``Verilogeval: Evaluating large language models for verilog code generation,'' \emph{arXiv preprint arXiv:2309.07544}, 2023.

\bibitem{corno2000rt}
F.~Corno, M.~S. Reorda, and G.~Squillero, ``{RT-level ITC'99 benchmarks and first ATPG results},'' \emph{IEEE Design \& Test of Computers}, 2000.

\bibitem{URL:opencore}
\emph{OpenCores: The reference community for Free and Open Source gateware IP cores}, https://opencores.org/.

\bibitem{amid2020chipyard}
A.~Amid, D.~Biancolin, A.~Gonzalez, D.~Grubb, S.~Karandikar, H.~Liew, A.~Magyar, H.~Mao, A.~Ou, N.~Pemberton \emph{et~al.}, ``Chipyard: Integrated design, simulation, and implementation framework for custom {SoCs},'' \emph{IEEE Micro}, 2020.

\bibitem{vexriscv}
\BIBentryALTinterwordspacing
VexRiscv, ``{VexRiscv: A FPGA friendly 32 bit RISC-V CPU implementation},'' 2022. [Online]. Available: \url{https://github.com/SpinalHDL/VexRiscv}
\BIBentrySTDinterwordspacing

\bibitem{wolf2013yosys}
C.~Wolf, J.~Glaser, and J.~Kepler, ``Yosys-a free verilog synthesis suite,'' in \emph{Austrian Workshop on Microelectronics (Austrochip)}, 2013.

\bibitem{guo2024deepseek}
D.~Guo, Q.~Zhu, D.~Yang, Z.~Xie, K.~Dong, W.~Zhang, G.~Chen, X.~Bi, Y.~Wu, Y.~Li \emph{et~al.}, ``Deepseek-coder: When the large language model meets programming--the rise of code intelligence,'' \emph{arXiv preprint arXiv:2401.14196}, 2024.

\bibitem{liu2024deepseek}
A.~Liu, B.~Feng, B.~Xue, B.~Wang, B.~Wu, C.~Lu, C.~Zhao, C.~Deng, C.~Zhang, C.~Ruan \emph{et~al.}, ``Deepseek-v3 technical report,'' \emph{arXiv preprint arXiv:2412.19437}, 2024.

\bibitem{liu2024visual}
H.~Liu, C.~Li, Q.~Wu, and Y.~J. Lee, ``Visual instruction tuning,'' \emph{Advances in neural information processing systems}, vol.~36, 2024.

\bibitem{li2023blip}
J.~Li, D.~Li, S.~Savarese, and S.~Hoi, ``Blip-2: Bootstrapping language-image pre-training with frozen image encoders and large language models,'' in \emph{ICML}, 2023.

\bibitem{li2022blip}
J.~Li, D.~Li, C.~Xiong, and S.~Hoi, ``Blip: Bootstrapping language-image pre-training for unified vision-language understanding and generation,'' in \emph{International Conference on Machine Learning (ICML)}, 2022.

\end{thebibliography}

\end{document}